\documentclass{article}

\usepackage[nonatbib,preprint]{neurips_2020}

\usepackage[utf8]{inputenc} % allow utf-8 input
\usepackage[T1]{fontenc}    % use 8-bit T1 fonts
\usepackage{hyperref}       % hyperlinks
\usepackage{url}            % simple URL typesetting
\usepackage{booktabs}       % professional-quality tables
\usepackage{amsfonts}       % blackboard math symbols
\usepackage{nicefrac}       % compact symbols for 1/2, etc.
\usepackage{microtype}      % microtypography
\usepackage{amsmath}
\usepackage{amssymb}
\usepackage{graphicx}
\usepackage{graphics}
\usepackage{mathtools}
\usepackage{wrapfig}

\title{Improving VQA and its Explanations \\ by Comparing Competing Explanations}

\author{%
  Jialin Wu\\
  Department of Computer Science \\
  University of Texas at Austin \\
  \texttt{jialinwu@cs.utexas.edu} \\
   \And
  Liyan Chen \\
    Department of Computer Science \\
  University of Texas at Austin \\
  \texttt{liyanc@cs.utexas.edu} \\
  \AND
  Raymond J. Mooney \\
    Department of Computer Science \\
  University of Texas at Austin \\
  \texttt{mooney@cs.utexas.edu}}

\begin{document}

\maketitle

\begin{abstract}
Most recent state-of-the-art Visual Question Answering (VQA) systems are opaque black boxes that are only trained to fit the answer distribution given the question and visual content. 
As a result, these systems frequently take shortcuts, focusing on simple visual concepts or question priors. This phenomenon becomes more problematic as the questions become complex that requires more reasoning and commonsense knowledge. To address this issue, we present a novel framework that uses explanations for competing answers to help VQA systems select the correct answer. By training on human textual explanations, our framework builds better representations for the questions and visual content, and then reweights confidences in the answer candidates using either generated or retrieved explanations from the training set. 
We evaluate our framework on the VQA-X dataset, which has more difficult questions with human explanations, achieving new state-of-the-art results on both VQA and its explanations.
\end{abstract}

\section{Introduction}

Recently, Visual Question Answering (VQA) \cite{antol2015vqa,hudson2019gqa,singh2019towards,marino2019ok} has emerged as a challenging task that requires artificial intelligence systems to predict answers by jointly analyzing both natural language questions and visual content. Most state-of-the-art VQA systems \cite{anderson2017bottom,kim2018bilinear,ben2017mutan,jiang2018pythia,cadene2019murel,lu2019vilbert,liu2019learning,tan2019lxmert} are trained to simply fit the answer distribution using question and visual features and achieve high performance on simple visual questions. However, these systems often exhibit poor explanatory capabilities and take shortcuts by only focusing on simple visual concepts or question priors instead of finding the right answer for the right reasons \cite{ross2017right,selvaraju2019taking}. This problem becomes increasingly severe when the questions require more complex reasoning and commonsense knowledge.

For more complex questions, VQA systems need to be right for the right reasons in order to generalize well to test problems. Two ways to provide these reasons are to crowdsource human visual explanations \cite{das2017human} or textual explanations \cite{park2018multimodal}. While visual explanations only annotate which parts of an image contribute most to the answer, textual explanations encode richer information such as detailed attributes, relationships, or commonsense knowledge that is not necessarily directly found in the image. Therefore, we adopt textual explanations to guide VQA systems.

\begin{wrapfigure}{l}{0.5\textwidth}
\includegraphics[width=\linewidth,trim={0cm 3.2cm 23.3cm 0cm},clip]{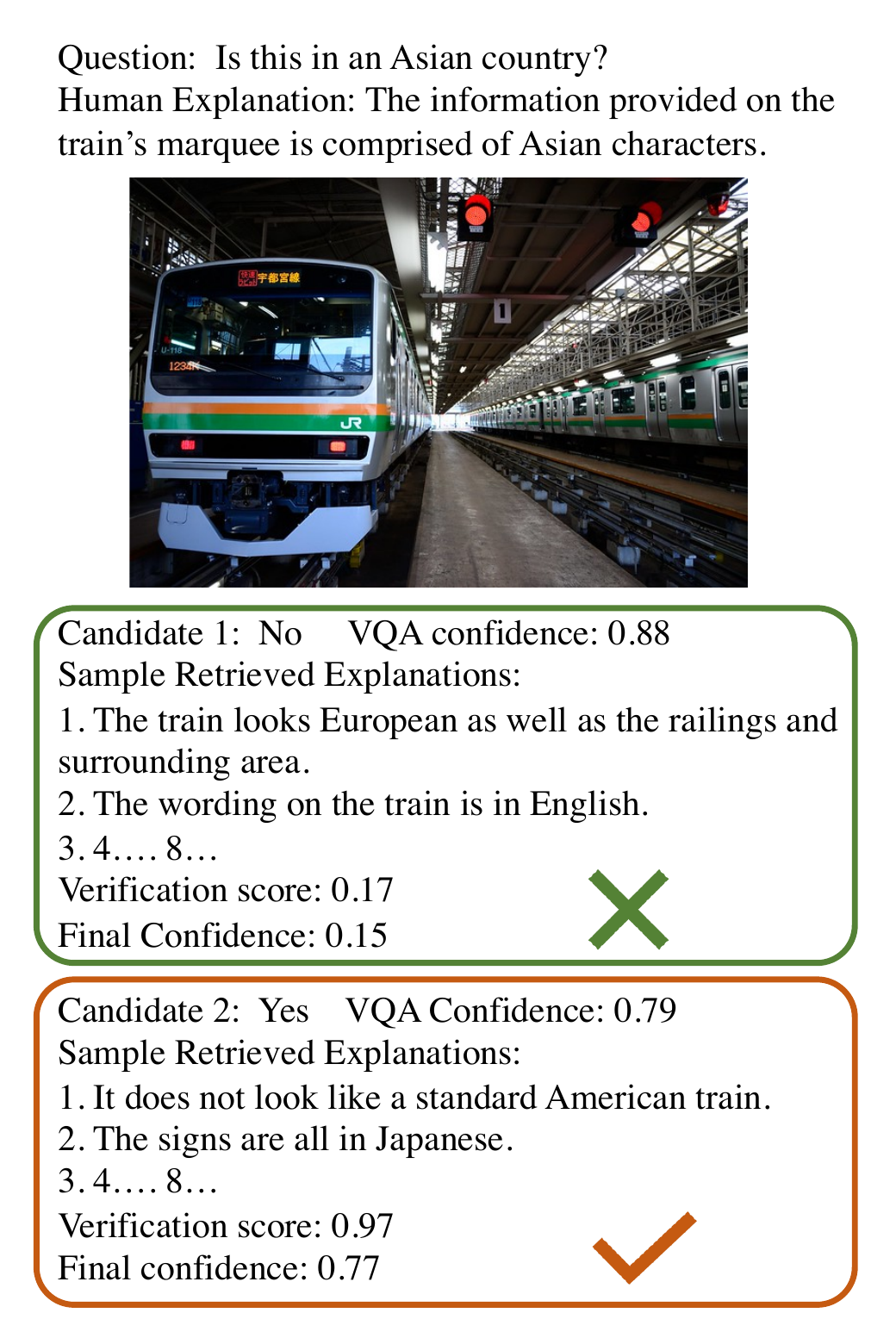}
\caption{An example of utilizing retrieved explanations to correct the original VQA prediction. Though the original VQA confidence of the correct answer ``Yes'' is lower than that of the incorrect answer ``No'', the retrieved explanations for ``Yes'' support their answer better, resulting in a higher verification score and a final correct decision.}
\vspace{-30pt}
\label{fig:demo}
\end{wrapfigure}

Recent research utilizing textual explanations adopts a multi-task learning strategy that jointly trains an answer predictor and an explanation generator \cite{li2018vqa, park2018multimodal}. However, this approach only considers explanations for the one chosen answer.  Our approach considers explanations for multiple competing answers, comparing these explanations when choosing a final answer, as shown in Figure \ref{fig:demo}.

Our framework is end-to-end trainable and therefore can be applied to any differentiable VQA system. Our experiments show improvements for our method combined with Up-Down \cite{anderson2017bottom} and LXMERT \cite{tan2019lxmert} on the VQA-X dataset \cite{park2018multimodal},  consisting of more complex questions estimated to require the abilities of at least a 9-year old and comes with human textual explanations. We also show that our approach learns better representations for the questions and visual content by training to retrieve explanations, and achieves state-of-the-art results by further jointly considering competing explanations.

We also developed a new VQA explanation generation approach utilizing competing explanations.  This approach uses retrieved explanations for each competing answer to help generate improved explanations during testing. Using both automated metrics comparing to human explanations and human evaluation, we show that these explanations are also improved, beating the current state-of-the-art method presented by \cite{wu2018faithful}.

\section{Related Work}
\subsection{VQA with Human Visual Explanations}
To train a VQA system to be right for the right reason, recent research has collected human visual attention \cite{das2017human,Gan_2017_ICCV} highlighting image regions that most contribute to the answer. Two popular approaches are to have crowdsourced workers deblur the image \cite{das2017human} or select segmented objects from the image \cite{Gan_2017_ICCV}. Then, the VQA systems try to align either the VQA system's attention \cite{qiao2018exploring,zhang2019interpretable} or the gradient-based visual explanation \cite{selvaraju2019taking,wu2019self} to the human attentions. These approaches help the systems focus on the right regions, and improve VQA performance when the training and test distributions are very different, such as in the VQA-CP dataset \cite{vqa-cp}.

\subsection{Human Textual Explanations}
While human visual explanations can help VQA systems know \textit{where} to attend, human textual explanations can also provide information on \textit{how} the attended image regions contribute to the answer. There are two textual explanation datasets,  VQA-E \cite{li2018vqa} and VQA-X \cite{park2018multimodal}. Explanations in VQA-E are automatically refined versions of the most relevant captions from the COCO dataset \cite{chen2015microsoft}, which have a larger scale but are of less quality. Therefore, we adopt VQA-X, where crowdsourced human workers were directly asked to provide textual explanations for the questions that are judged to require children older than 9 years to answer.

\subsection{Generating Textual Explanations}
In order to automatically generate textual explanations, \cite{park2018multimodal} present a single-layer LSTM network trained on image, question, and answer features to mimic crowdsourced human explanations. \cite{wu2018faithful} uses question-attended segmentation features from the original VQA system as input, and try to generate explanations that are more faithful to the actual VQA process at the object level.

\subsection{VQA with Human Textual Explanations}
For more complex questions requiring reasoning and general knowledge, visual attention, which only shows important regions, is less helpful. We know of only two papers that use textual explanations to aid VQA. \cite{li2018vqa} train to jointly predict the answer and generate an explanation. However, the explanations may not faithfully reflect the actual VQA process, can hallucinate visual content \cite{rohrbach2018object}, and therefore may not properly supervise the underlying VQA system. \cite{wu2019self} only use textual explanations to extract a set of important visual objects, but ignore other critical richer content, $e.g.,$ attributes, relationships, commonsense knowledge, etc. In contrast, our approach trains a system to distinguish correct human explanations from competing explanations supporting incorrect answers.

\section{Baseline VQA Models}

Many recent VQA systems \cite{fukui2016multimodal,ben2017mutan,ramakrishnan2018overcoming} utilize a trainable top-down attention mechanism over convolutional features to recognize relevant image regions. Up-Down (UpDn) \cite{anderson2017bottom} introduced complementary bottom-up attention that first detects common objects and attributes so that the top-down attention can directly model the contribution of higher-level concepts. Several recent systems have used this approach \cite{selvaraju2019taking,jiang2018pythia,lu2019vilbert,tan2019lxmert}, significantly improving VQA performance. These systems first extract a visual feature set $\mathcal{V}$ = $\{\textbf{v}_i, ..., \textbf{v}_{|\mathcal{V}|}\}$ for each image whose element $\textbf{v}_i$ is a feature vector for the $i$-th detected object. On the language side, UpDn systems sequentially encode each question $Q$ to produce a question vector $\textbf{q}$. Let $f$ denote the answer prediction operator that takes both visual features and question features as input and predicts the confidence for each answer $a$ in the answer candidate set $\mathcal{A}$, $i.e.$  $P(a|\mathcal{V}, Q) = f(\mathcal{V}, \textbf{q})$. The VQA task is framed as a multi-label regression problem with the gold-standard soft scores as targets in order to be consistent with the evaluation metric. Finally, binary cross entropy loss with soft score is used to supervise the sigmoid-normalized outputs.

We briefly introduce two variants of this approach adopted in our experiments:

\noindent\textbf{UpDn}. This is the original UpDn system, which uses a single layer GRU to encode questions. The question vector is then used to compute a single-stage attention over the detected objects to produce attended visual features. Finally, a two-layer feed-forward network computes answer probabilities given the joint features of the question and visual content . 

\noindent\textbf{LXMERT}. In order to learn richer representations for both questions and visual content, LXMERT \cite{tan2019lxmert} uses transformers \cite{vaswani2017attention,devlin2018bert} that learn multiple layers of attention over the input. In particular, it first learns 9 layers over the input question and 5 layers over detected objects, then finally learns another 5 layers of attention across the two modalities to produce the final joint representation.

\section{Approach}
\begin{figure*}[]
    \centering
    \includegraphics[width=\linewidth,trim={0cm 11.cm 18cm 0cm},clip]{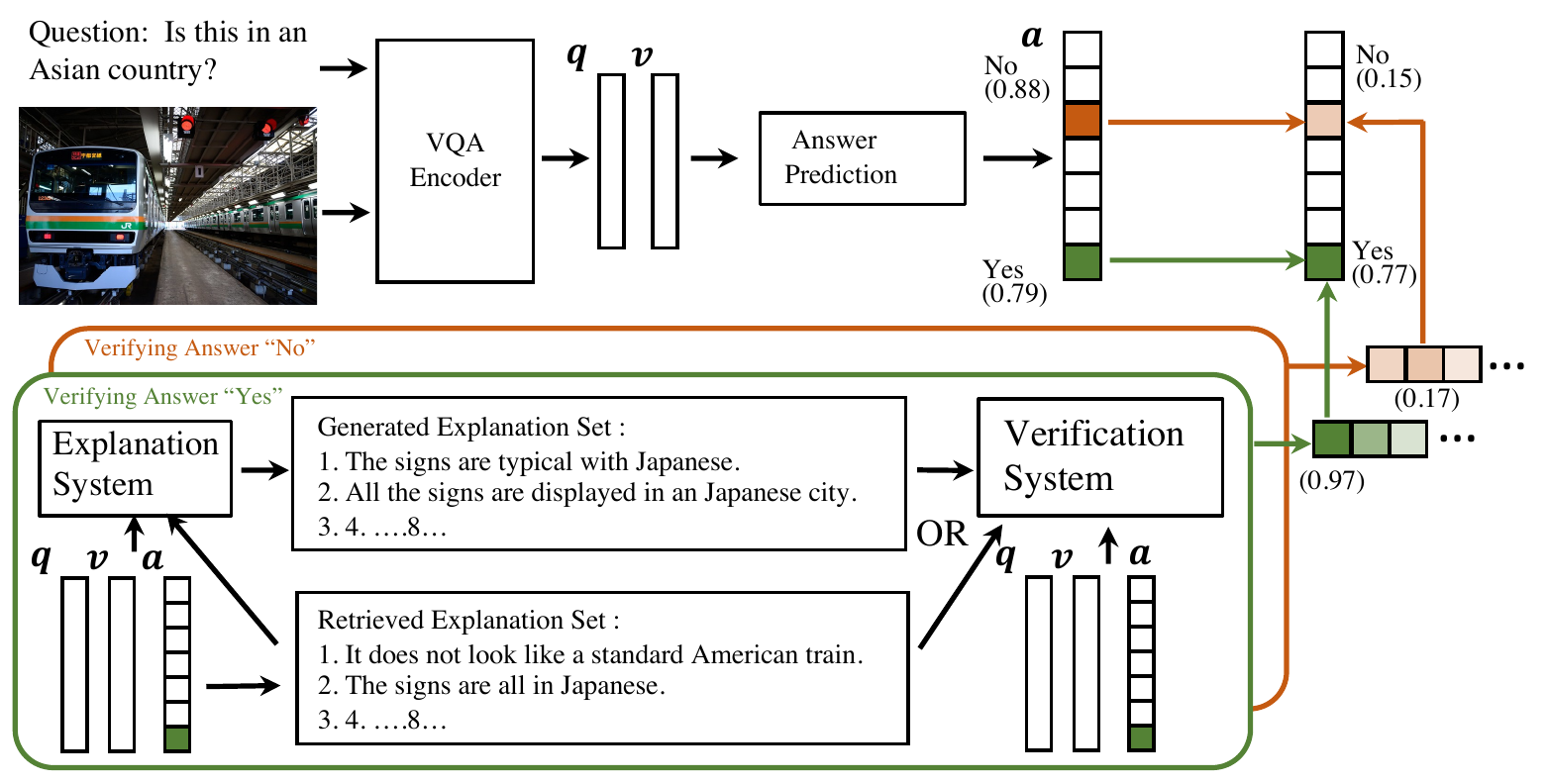}
    \caption{Our approach first predicts a set of answer candidates and retrieves explanations for each based on the answer, question, and visual content. These explanations are then used to generate improved explanations. Finally, either retrieved or generated explanations are employed to predict verification scores that are used to reweight the original predictions and compute the final answer.}
    \label{fig:model}
\end{figure*}

This section presents our approach to utilizing competing explanations to aid VQA. As shown in Figure \ref{fig:model}, after the base VQA system computes the top-$k$ answers, our approach retrieves the most supportive explanations for each answer from the training set to construct the set of competing explanations. Then, these explanations are used to help generate explanations for the current question. Next, we learn to predict verification scores that indicates how well the retrieved or generated explanations support the predictions given the input question and visual content. The final answer is determined by jointly considering the original answer probabilities and these verification scores.
\subsection{Retrieving Explanations}
This section presents our approach to retrieving the most supportive human textual explanation from the training set for each answer candidate. Ideally, we should dynamically retrieve explanations for each answer at each iteration. However, it will be very computational costly because the question and visual features have to be computed for each image from the training set. Therefore, we adopt the below relaxation for computational efficiency that only needs to compute the features once.

In particular, we first pretrain the VQA system, and extract the question and visual embeddings, $\textbf{q}$ and $\textbf{v}$ ,for each $Q\mathcal{V}$ pair in the training set. For UpDn, we use the attended visual features and the question GRU's last hidden state as the visual and question embeddings. For LXMERT, we use the last cross-modal attention layer's visual and question output as the embeddings.

Then, for each $Q\mathcal{V}$ pair, we only compute the top-10 answer candidates since the top-10 answers have already achieved high recall. After that, for each answer candidate $a$, we extract explanations from the training set that have the same ground truth answer \footnote{More specifically, the soft score of the answer candidate in the retrieved explanation's example is over 0.6} as the current candidate. We then sort these explanations by the L2 distance between the explanations' $Q\mathcal{V}$ embeddings, $\textbf{q} \odot \textbf{v}$, and the example's and pick the closest 8 explanations as the competing explanations set denoted as $\mathcal{X}_a$. 

\subsection{Generating Explanations}
Next,the retrieved explanations for similar VQA examples from the training set are used to help generate even better explanations.

We adopt the explainer from \cite{wu2018faithful}, a two-layer LSTM network similar to the UpDn captioner \cite{anderson2017bottom}, as our baseline. Since the current VQA systems are built upon detected objects, we use them as the visual inputs instead of segmentations. 

The baseline explainer first computes a set of question-attended visual features, $\mathcal{U}$, and an average pooled version, $\bar{\textbf{u}}$.  The explainer then uses $\bar{\textbf{u}}$ and $\mathcal{U}$ together with question and answer embeddings as inputs to produce explanations.

Our approach simply replaces the average pooled question-attended visual features $\bar{\textbf{u}}$ with the retrieved explanations' features, $\textbf{x}$. We use a single-layer GRU to encode all of the retrieved explanations for the correct answer, and then max pool the last hidden states among these explanations to compute $\textbf{x}$.
We sample 8 explanations for each answer candidate to construct the generated explanation set.

\subsection{Learning Verification Scores}
Next, a verification system is trained to score how well a generated or retrieved explanation supports a corresponding answer candidate given the question and visual content. The verification system takes four inputs: the visual, question, answer and its explanation features; and outputs the verification score, $i.e.$ $S(Q, \mathcal{V}, a, x) = \sigma (f_2(f(\textbf{q}), f(\textbf{v}), f(\textbf{a}), f(\phi (x)))$.

where $\textbf{a}$ is the one-hot embedding of the answer, and $\phi(x)$ is the feature vector for the explanation, $x$, encoded using a GRU \cite{cho2014learning}, $\phi$. We use $f_n$ to denote $n$ consecutive feed-forward layers (for simplicity $n$ is omitted when $n=1$). We use $\sigma$ to denote the sigmoid function. The verification system is similar to the answer predictor in architecture except for the number of outputs, $i.e.$ 1 for the verification system and $3,129$ for the answer predictor.

Given the VQA examples with their explanations in the VQA-X dataset, we use binary cross-entropy loss $\mathcal{L}_m$ to maximize the verification score for the matching human explanations, $i.e.$ $\mathcal{L}_m = -\log(S(Q, \mathcal{V}, a, x)) \label{eq:verif_match}$.

Intuitively, we want the verification score $S$ to be high only when the explanation is matched to the VQA example, $i.e.$ replacement of any of the four input sources should lower the score. Therefore, we designed the five kinds of replacements below for constructing negative examples.

\noindent \textbf{Replacement of Visual and Question Features:}
Ideally, we should replace the visual and question features with the complementary features \cite{antol2015vqa} that lead to the opposite answer. For example, for the question ``Is this a vegetarian pizza?'', with an image of a vegetarian pizza, we should replace the image with one of a meat pizza, $i.e. $ counter-factual images. However, such replacement requires retrieving and computing the visual features $\textbf{v}$ for the meat pizza, which is computationally inefficient. Therefore, we simply randomly choose a $Q'$ or $\mathcal{V}'$ replacement from the current batch and minimize the binary cross entropy loss $\mathcal{L}_r^q$, $\mathcal{L}_r^q$ for the verification scores, $i.e.$ $\mathcal{L}_r^q = -\log( 1 - S(Q', \mathcal{V}, a, x)) , ~  \mathcal{L}_r^v =  -\log( 1 -S(Q, \mathcal{V}', a, x))$.

\noindent \textbf{Replacement of Answer Features:}
We sample the answer for replacement according to the current VQA's predicted incorrect probabilities. At each step, we try to minimize the expectation of binary cross-entropy loss  $\mathcal{L}_r^a$ for the incorrect predictions,
$i.e.$ $\mathcal{L}_r^a =  \mathbb{E}_{a' \sim p(a'| QV), ~s(a') < 0.6 } [ -\log( 1 - S(Q, \mathcal{V}, a', x)) ] $
where $s(a')$ denotes the human VQA soft score for answer $a'$. In practice, we only sample one incorrect answer during training.

\noindent \textbf{Replacement of Explanation Features:}
We try to replace the matched human explanations with the most supportive explanations for the sampled incorrect answer and train our verification system to disprefer that explanation using the loss $\mathcal{L}_r^a = \max_{x' \in \mathcal{X}_{a'}} [ -\log( 1 - S(Q, \mathcal{V}, a, x'))] $. In particular, given the sampled incorrect answer $a'$ from the previous section, we compute the verification score for each retrieved or generated  explanation $x'$ from the set $\mathcal{X}_{a'}$ for that wrong answer and regard the one with maximum verification score as the most supportive one. 

\noindent \textbf{Replacement of Answer and Explanation Features:}
To further prevent the system from being falsely confident in the sampled incorrect answer $a'$, we also minimize the verification score for its most supportive explanation for the incorrect answer $a'$, $i.e.$ $\mathcal{L}_r^{ax} = -\log (\max_{x' \in \mathcal{X}_{a'}} (1 - S(Q, \mathcal{V}, a', x')))$

To sum up, the verification loss is the sum of the aforementioned 6 losses as shown in Eq. \ref{eq:verif_loss}: 
\begin{align}
   \mathcal{L}_{verification} =  \lambda \mathcal{L}_m + \mathcal{L}_r^{q}+ \mathcal{L}_r^{v} + \mathcal{L}_r^{a} + \mathcal{L}_r^{x} +\mathcal{L}_r^{ax} \label{eq:verif_loss}
\end{align}
Since we have more negative examples and only one positive examples, we assign a larger loss weight ($i.e.$ $\lambda=10$) for the only positive example.

\subsection{Using Verification Scores}
The original VQA system provides the answer probabilities conditioned on the question and visual content, $i.e.$ $P(a|Q, \mathcal{V})$. The verification scores $S(Q, \mathcal{V}, a, x)$ are further used to reweight the original VQA predictions so that the final predictions $\tilde{P}(a|Q, \mathcal{V})$, shown in Eq. \ref{eq:chain}, can take the explanations into account.
\begin{align}
    \tilde{P}(a|Q, \mathcal{V}) =  P(a|Q, \mathcal{V}) \max_{x \in \mathcal{X}_a} S(Q, \mathcal{V}, a, x) \label{eq:chain}
\end{align}
where the $\mathcal{X}_a$ denotes the generated or retrieved explanation set for the answer $a$.

Since we try to select the correct answer with its explanation, the prediction $\tilde{P}(a|Q, \mathcal{V})$ should only be high when the answer $a$ is correct and the explanation $x$ supports $a$, which is enforced using the loss in Eq. \ref{eq:vqa_loss}:
\begin{align}
\mathcal{L}_{vqae} =   -\log(P(a|Q, \mathcal{V}) S(Q, \mathcal{V}, a, x_a) )  -\log( 1 -  \tilde{P}(a'|Q, \mathcal{V})) \label{eq:vqa_loss}
\end{align}
where the $x_a$ denotes the human explanation for the answer $a$.

During testing, we first extract the top 10 answer candidates $\mathcal{A}$, and then select the explanation for the answer candidate with the highest verification score. Then, we compute the explanation-reweighted score for each answer candidate to determine the final answer $
    a^\star = \arg\max_{a \in \mathcal{A}}  \tilde{P}(a|Q, \mathcal{V})$.

\iffalse
\begin{table*}[!t]
\centering
\begin{tabular}{l|c|cccc}
\hline \toprule
                    & Training set & \multicolumn{4}{c}{VQA-X } \\    \hline
                    &       &   Overall     & Yes/No &  Num  & Other \\ \hline\hline
UpDn    \cite{anderson2017bottom}   &VQA-X & 74.2   &  70.9  & 5.0  & 77.0  \\ 
UpDn+E (ours)   &  VQA-X  &  78.7   &  78.5  &  5.0 &  79.1 \\\hline
LXMERT \cite{tan2019lxmert}   &  VQA-X  &  76.8   &  75.4  &  0.0 & 78.2   \\
LXMERT+E (ours)   &  VQA-X  &  78.0   &  76.0  & 0.0  & 79.8  \\  \hline\hline
UpDn    \cite{anderson2017bottom}   & VQA-v2 & 83.6   &  82.2  & 0.0  & 85.2  \\ 
UpDn+E (ours)   &  VQA v2  &  85.4   &  84.7  & 0.0  & 86.3  \\\hline
LXMERT  \cite{tan2019lxmert}  &  VQA v2  &  83.7   &  82.5  & 5.0  & 85.0  \\
LXMERT+E (ours)   &  VQA v2  &  84.7   &  84.0  & 0.0 & 85.7  \\ \bottomrule
\end{tabular}
\caption{Experimental results comparison on VQA-X dataset. The first and second half reports the results for the VQA systems trained on VQA-X and VQA v2 training set respectively. ``+E'' denotes using our competing explanations approach.}
\label{tab:crit_ablation}
\end{table*}
\fi

\subsection{Training and Implementation Details}

\noindent\textbf{Training Details.} We first pre-train our base VQA system (Up-Down or LXMERT) on either the entire VQA v2 training set for 20 epochs or only the VQA-X training set for 30 epochs with the standard VQA loss (binary cross-entropy loss with soft scores as supervision) and the Adam optimizer \cite{kingma2014adam}. As the VQA-X validation and test set are both from the VQA v2 validation set that is covered in the LXMERT pretraining, we do not use the officially released LXMERT parameters. The learning rate is fixed to 5e-4 for UpDn and 5e-5 for LXMERT, with a batch size of 384 during the pre-training process. For answer prediction part, We use $1,280$ hidden units in UpDn and $768$ hidden units in LXMERT, and for verification part, we use $1,280$ hidden units in both systems.

We fine-tune our system using the verification loss and VQA loss $\mathcal{L}_{verification} + 0.1\mathcal{L}_{vqae}$ on the VQA-X training set for another 40 epochs. The initial learning rate for VQA system is set to be same as the pretraining if the system is pretrained on VQA-X training set, and 0.1$\times$ if the system is pretrained on VQA v2 training set. For the verification systems, the initial learning rate is set to 0.0005. The learning rate for every parameter is decayed by 0.8 every 5 epochs.
During test, we consider the top-10 answer candidates for the VQA systems and use the explanation-reweighted prediction as the final answer.

\noindent\textbf{Implementation.}
We implemented our approach on top of the original UpDn and LXMERT. Both base systems utilize a Faster R-CNN head \cite{girshick2015fast} in conjunction with a ResNet-101 base network \cite{he2016deep} as the object detection module. The detection head is pre-trained on the Visual Genome dataset \cite{krishna2017visual} and is capable of detecting $1,600$ objects categories and $400$ attributes. Both base systems take the final detection outputs and perform non-maximum suppression (NMS) for each object category using an IoU threshold of $0.7$.  Convolutional features for the top $36$ objects are then extracted for each image as the visual features, $i.e.$ a $2,048$ dimensional vector for each object. For question embedding, following \cite{anderson2017bottom}, we perform standard text pre-processing and tokenization for UpDn. In particular, questions are first converted to lower case, trimmed to a maximum of $14$ words, and tokenized by white spaces. A single layer GRU \cite{cho2014learning} is used to sequentially process the word vectors and produce a sentential representation for the pre-processed question. We also use Glove vectors \cite{pennington2014glove} to initialize the word embedding matrix when embedding the questions. 
For LXMERT, we also follow the original BERT word-level sentence embedding strategy that first splits the sentence into words $w_1,...,w_n$ with length of $n$ by the same WordPiece tokenizer \cite{wu2016google} in \cite{devlin2018bert}. Next, the word and its index ($i.e.$ absolute position in the sentence) are projected to vectors by embedding sub-layers,  and then added to the index-aware word embeddings. 
We use a single-layer GRU and three-layer GRU to encode the generated or retrieved explanation in the verification system when using UpDn and LXMERT as base system, respectively.

\begin{table*}[!t]
\centering
\begin{tabular}{l|cc|cc}
\hline \toprule
                    & \multicolumn{2}{c|}{VQA-X Pretrain} & \multicolumn{2}{c}{VQA v2 Pretrain} \\    \hline
                    &     Gen. Expl.     & Ret. Expl. &  Gen. Expl.  & Ret. Expl. \\ \hline\hline
UpDn    \cite{anderson2017bottom}    & 74.2   &  74.2  & 83.6  & 83.6  \\ 
UpDn+E (ours)   &  78.0   &  78.7  &  85.1 &  85.4 \\\hline
LXMERT \cite{tan2019lxmert}   &  76.8   &  76.8  &  83.7 & 83.7   \\
LXMERT+E (ours)   &  77.3   &  78.0  & 84.1  & 84.7  \\  \bottomrule
\end{tabular}
\caption{Question answering accuracy on VQA-X using both UpDn and LXMERT as a base system,``+E'' denotes using our competing explanations approach. ``Gen. Expl.'' and ``Ret. Expl.'' denote using generated and retrieved explanations, respectively.}
\label{tab:crit_ablation}
\end{table*}

\begin{table*}[t]
\centering
\begin{tabular}{l|ccccc|c}
\toprule
        &   \multicolumn{5}{c|}{Automatic Evaluation}& AMT   \\\hline
        & BLEU-4 & METEOR & ROUGE & CIDEr & SPICE        &               \\ \hline\hline
Faith. Expl. \cite{wu2018faithful}&  25.0   & 20.0   & 47.1  & 91.1  & 18.6 & 49.5 \\
Faith. Expl. + E (ours) &  26.4   & 20.4   & 48.5  & 95.3  & 18.7 & 55.6 \\\bottomrule               
\end{tabular}
\caption{Automatic and human evaluation of our generated explanations. ``AMT'' denotes the human evaluation scores.}
\label{tab:results}
\end{table*}

\section{Experimental Results}
This section presents experimental results on the VQA-X \cite{park2018multimodal} dataset where the questions require more cognitive maturity than the original VQA-v2 dataset. We combine the validation set (1,459 examples) and test set (1,968 examples) of the VQA-X dataset as our larger test set (3,427 examples) for more stable results since both  are relatively small. We compare our system's VQA performance against two corresponding base systems using the standard protocol. In addition, we examine the quality of explanations by comparing our system against a baseline model as well as human explanations. Finally, we perform ablation studies to show that both improved feature representation and explanation reweighting  are key aspects of the improvements. 

Also, we report the ablation results on each way of constructing negative examples and qualitative examples in the supplementary materials.

\subsection{Results on VQA performance}
Table \ref{tab:crit_ablation} reports the results of our competing explanation approach. Our approach combined with UpDn pretrained on the entire VQA v2 dataset achieves the best results. When training only on the VQA-X training set, we improve the original UpDn and LXMERT by 4.5 \% and 1.2 \%, respectively. UpDn benefits more from using competing explanations than LXMERT, but both improve.   By using transformers, LXMERT already creates better, but less flexible, representations which are harder to improve upon by using explanations.

\subsection{Results on Explanation Generation}
We evaluated the generated explanations using both automatic evaluation metrics comparing them to human explanations ($i.e.$ BLEU-4 \cite{Papineni:2002:BMA:1073083.1073135}, METEOR \cite{banerjee2005meteor}, ROUGE-L \cite{lin2004rouge}, CIDEr \cite{vedantam2015cider} and SPICE \cite{spice2016}) and human evaluation using Amazon Mechanical Turk (AMT) platform. The explanation generator uses the features from the UpDn VQA system pretrained on VQA v2 and employs beam search with a beam size of two.  We compare to a recent state of the art VQA explanation system \cite{wu2018faithful} as a baseline system.

In the AMT evaluation, we ask human judges to compare generated explanations to human ones. We randomly sampled 500 such pairs of explanations for both the baseline and our system. In order to measure the difference in explanation quality relative to standard oracles, we perform two groups of comparisons: our system v.s. oracles and baseline v.s. oracles, where each group contains $500$ randomly sampled sets of comparisons.
Each comparison consists of a (question, image, answer) triplet and two randomly ordered explanations, one human and one generated. Three turkers were asked which explanation was a better justification of the answer, or declare a tie. We aggregated results by assigning 2 points to winning explanations, 0 to losing ones, and 1 to each in the case of a tie, such that each comparison is zero-sum and robust to noise to some extent.
%Each AMT HIT (Human Inference Task) contains 3 tasks. 
We normalized the total score for automated explainers by the score for human explanations, so the final score ranges from 0 to 1, where 1 represents equivalent to human performance.

As reported in Table \ref{tab:results}, by additionally conditioning on human explanations for similar visual questions, the generated explanations achieve both better automatic scores, and more importantly, higher human ratings.

\subsection{Effect of Using Different Explanations}

This section compares variations of our approach using different sources of explanations, including generated, retrieved and human ones. Table \ref{tab:ablation_expl} reports overall VQA scores using UpDn pretrained on the VQA-X train set. We include two baseline settings, "UpDn" and "UpDn + VQA-E", where the model is trained to jointly predict the answer and generate the explanation, using a two-layer attentional LSTM on top of the VQA shared features. This version models the approach used in \cite{li2018vqa}.

In the first human explanation setting, denoted by $\mathcal{RR}$, we only $\mathcal{R}$eplace the retrieved explanation for the $\mathcal{R}$ight answer with the corresponding human ones, and still use retrieved explanation for the incorrect answer. This setting shows how much retrieved explanations for correct answers impacts the results. The second human explanation setting, denoted by $\mathcal{RA}$, assumes that human explanations are used to $\mathcal{R}$eplace the retrieved explanations for $\mathcal{A}$ll the potential answer candidates. This setting provides an upper bound on our approach that uses textual explanations.

\begin{table}[t]
    \begin{minipage}{.5\linewidth}
      \centering
        \begin{tabular}{l|c}
\hline \toprule
                  & VQA-X \\ \hline
UpDn \cite{anderson2017bottom} &  74.2 \\
UpDn + VQA-E \cite{li2018vqa} &   76.0 \\ \hline
UpDn + generated explanations  &   78.0 \\
UpDn + retrieved explanations &    78.7 \\
UpDn + human explanations ($\mathcal{RR}$) &   79.3 \\
UpDn + human explanations ($\mathcal{RA}$)&    80.2 \\\bottomrule
\end{tabular}
\caption{Comparison of VQA performance using different explanations.}
\label{tab:ablation_expl}
    \end{minipage}%
    \quad
\begin{minipage}{.5\linewidth}
      \centering
        \begin{tabular}{l|c}
\hline \toprule
                  & VQA-X \\ \hline
UpDn \cite{anderson2017bottom} &  74.2\\
UpDn + VQA-E \cite{li2018vqa} &   76.0 \\ \hline
UpDn+E ($w/o$ reweighting) &   77.8\\
UpDn+E (fixed VQA) &    75.3\\
UpDn+E  &   78.7\\\bottomrule
\end{tabular}
\caption{Ablation studies on representation improvements. }
\label{tab:repr}
    \end{minipage} 
\end{table}

The results indicate that using explanations even in a simple joint model \cite{li2018vqa} is helpful, providing 2\% improvement on the overall score. However,
our competing explanation approach with either generated or retrieved explanations significantly outperforms both baseline models.

Our system with retrieved explanations performs slightly better than the one with generated explanations. This is probably because there are no guarantees that the generated explanations will support the answers upon which they are conditioned. Since all the explanation training examples are for correct answers, the explanation generation system tends to support the ground truth answer regardless of the answer candidate it is generated to support. Also, the generated explanations sometimes ignore or hallucinate \cite{rohrbach2018object} visual content when explaining the answer. Therefore, although ideally, generated explanations could work better than retrieved ones, they are currently less helpful to the VQA performance due to their imperfections.

Not surprisingly, the human oracle explanations help the VQA system more than the retrieved ones.  It indicates that our approach could achieve even better performance  with more informative explanations, which could be achieved by either developing a better explanation generator, or enlarging the explanation training set from which human explanations are retrieved. Our system's results using retrieved explanations is only 0.6\% lower than with human oracle explanations for the correct answers. This indicates that the retrieved explanations (for related questions) are a reasonable approximation to human explanations for the specific question.

\iffalse
\subsection{Ablation Study on Each Kind of Replacement. }
Table \ref{tab:ablation} reports the results of ablating each one kind of replacement for raising negative examples during training the verification model. We use the UpDn model pretrained on VQA-X dataset. It verifies the value of each term in Eq. \ref{eq:verif}.

\begin{table*}[!t]
\centering
\begin{tabular}{l|c|c|c|c|c|c}
\hline \toprule
                    & Full& $w/o$ $q$& $w/o$ $v$& $w/o$ $a$& $w/o$ $x$&$w/o$ $ax$\\   \hline
UpDn+Re.Expl. (ours)  &78.7 &  77.2   &  77.4  & 77.9   & 78.4 & 78.3 \\\bottomrule
\end{tabular}
\caption{Ablation study on each kind of replacements.}
\label{tab:ablation}
\end{table*}
\fi

\subsection{Evaluating Representation Improvement}
This section presents an ablation investigating how our approach improves the learned representations. The ``$w/o$ reweighting'' ablation still uses the fine-tuned representation trained using explanations, but it does not reweight the final predictions, therefore it tests the improvement solely due to better joint representations for the question and the visual content. The ``fixed VQA'' ablation uses reweighting, but does not fine-tune the VQA parameters during verification-score training ($i.e.$ only the verification parameters are trained).

Table \ref{tab:repr} reports the results of the UpDn system pretrained on VQA-X dataset. Using explanations as additional supervision helps the VQA systems are able to build better representations for the question and answer, improving performance by 3.6\%. This is because minimizing the verification loss $\mathcal{L}_{verification}$ prevents the VQA system from taking shortcuts. First, the $\mathcal{L}_m$ component forces the VQA system to produce visual and question features whose mapping can match the explanation features. Second, by minimizing $\mathcal{L}_{r}^{v}$ and $\mathcal{L}_{r}^{v}$, the system is forced not to solely focus on question and/or visual priors.
Finally, our full system gains 1.1\% improvement due to reweighting, and achieves our best results.

%\begin{table}[h]
%\centering
%\begin{tabular}{l|c}
%\hline \toprule
%                  & VQA-X \\ \hline
%LXMERT \cite{anderson2017bottom} &  76.8\\
%LXMERT+E ($w/o$ reweight) &   77.4\\
%LXMERT+E (fixed VQA) &    -\\
%LXMERT+E  &    78.0\\\bottomrule
%\end{tabular}
%\caption{Comparison of the performance using generated and retrieved captions.}
%\label{tab:human_retrieve_compare}
%\end{table}

\section{Conclusion and Future Work}
In this work, we have explored how to improve VQA performance by comparing competing explanations for each answer candidate. We present two sets of competing explanations, generated and retrieved explanations. Our approach first helps the system learn better visual and question representations, and also reweight the original answer predictions based on the competing explanations. As a result, our VQA system avoids taking shortcuts and is able to handle difficult visual questions better, improving results on the challenging VQA-X dataset. We also show that our approach generates better textual explanations by additionally conditioning on the retrieved explanations for similar questions. In the future,  we would like to combine different sorts of explanations (e.g. both generated and retrieved ones) together to better train VQA systems. 

\section{Broader Impact}
\subsection{Positive} The focus of this work is to learn a more sophisticated VQA system to better answer complex visual questions that require more reasoning and commonsense knowledge. By using the competing explanations for each potential answer candidates, we encourage the system to use the same rationales as humans, preventing the system from taking short-cuts during inference by only focusing on answer priors and simple visual perceptual concepts. 

Meanwhile, our VQA system also associates each question with either retrieved or generated explanations to illustrate the rationale behind the answer, providing users with more insight and understanding to the system's reasoning that provide more transparency and help engender trust.

Hopefully, both increased accuracy and better explanation will help improve VQA systems for important socially-beneficial applications, including among others, aids for the visually impaired and improved analysis and interpretation of medical imagery.

\subsection{Negative}
Though the proposed systems attempt to generate ``faithful'' explanations that are biased to focus on aspects of the image that are explicitly attended to by the underlying VQA network, it also tries to mimic human explanations and may generate rationales that do not completely faithfully reflect all of the details of the system operation.
Therefore, the system might generate incorrect answers but reasonable explanations, which could potentially convince users to accept erroneous conclusions, which could have significant negative impacts. 

In addition, it is possible that the retrieved or generated explanations could introduce new undesirable biases ($e.g.$ gender bias) when reweighting the original predictions, particularly if such biases are present in the human explanations used to train the system. Careful vetting of the training explanations to prevent such biases is desirable but would be difficult and costly. 

%Also, the proposed algorithm requires collecting human explanations for training examples, which could be infeasible for some areas. For example, having experts and doctors annotate a large number of medical visual questions are too expensive. 

\bibliographystyle{ieee}
\bibliography{nips2020.bib}

\end{document}